\documentclass[10pt,twocolumn,letterpaper]{article}

\usepackage{wacv}
\usepackage{times}
\usepackage{epsfig}
\usepackage{graphicx}
\usepackage{amsmath}
\usepackage{amssymb}

\usepackage{color}          
\usepackage{microtype}      
\usepackage{dcolumn}        
\usepackage{hyperref}       
\usepackage{url}            
\usepackage{booktabs}       
\usepackage{amsfonts}       
\usepackage{nicefrac}       
\usepackage{multirow}
\usepackage{float}

\wacvfinalcopy 

\def\wacvPaperID{****} 

\ifwacvfinal\fi
\begin{document}

\title{Domain Randomization for Scene-Specific Car Detection and Pose Estimation}

\author{Rawal Khirodkar \hspace{2cm} Donghyun Yoo \hspace{2cm} Kris M. Kitani \\
Carnegie Mellon University\\
{\tt\small \{rkhirodk,donghyuy,kkitani\}@cs.cmu.edu}
}

\maketitle
\ifwacvfinal\thispagestyle{empty}\fi

\begin{abstract}

    We address the issue of domain gap when making use of synthetic data to train a scene-specific object detector and pose estimator. While previous works have shown that the constraints of learning a scene-specific model can be leveraged to create geometrically and photometrically consistent synthetic data, care must be taken to design synthetic content which is as close as possible to the real-world data distribution. In this work, we propose to solve domain gap through the use of appearance randomization to generate a wide range of synthetic objects to span the space of realistic images for training. An ablation study of our results is presented to delineate the individual contribution of different components in the randomization process. We evaluate our method on VIRAT, UA-DETRAC, EPFL-Car datasets, where we demonstrate that using scene specific domain randomized synthetic data is better than fine-tuning off-the-shelf models on limited real data.  
\end{abstract}

\section{Introduction}

\begin{figure}[t]
\centering
\begin{center}
\resizebox{3.5in}{!}{%
\begin{tabular}{c}
\includegraphics[width=0.95\linewidth]{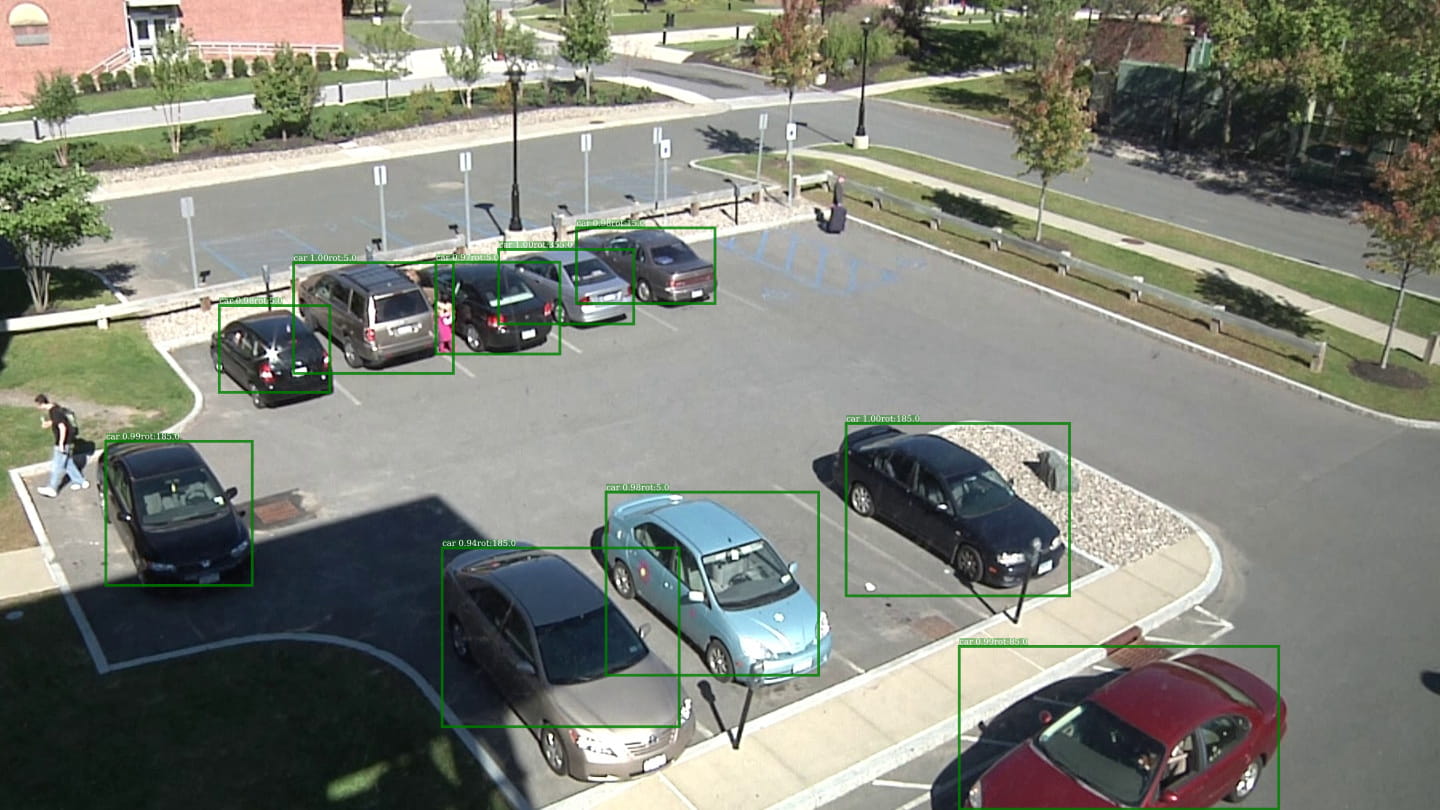}
\label{fig:bb_det_intro}
\\
(a) Car detection
\\
\includegraphics[width=0.95\linewidth]{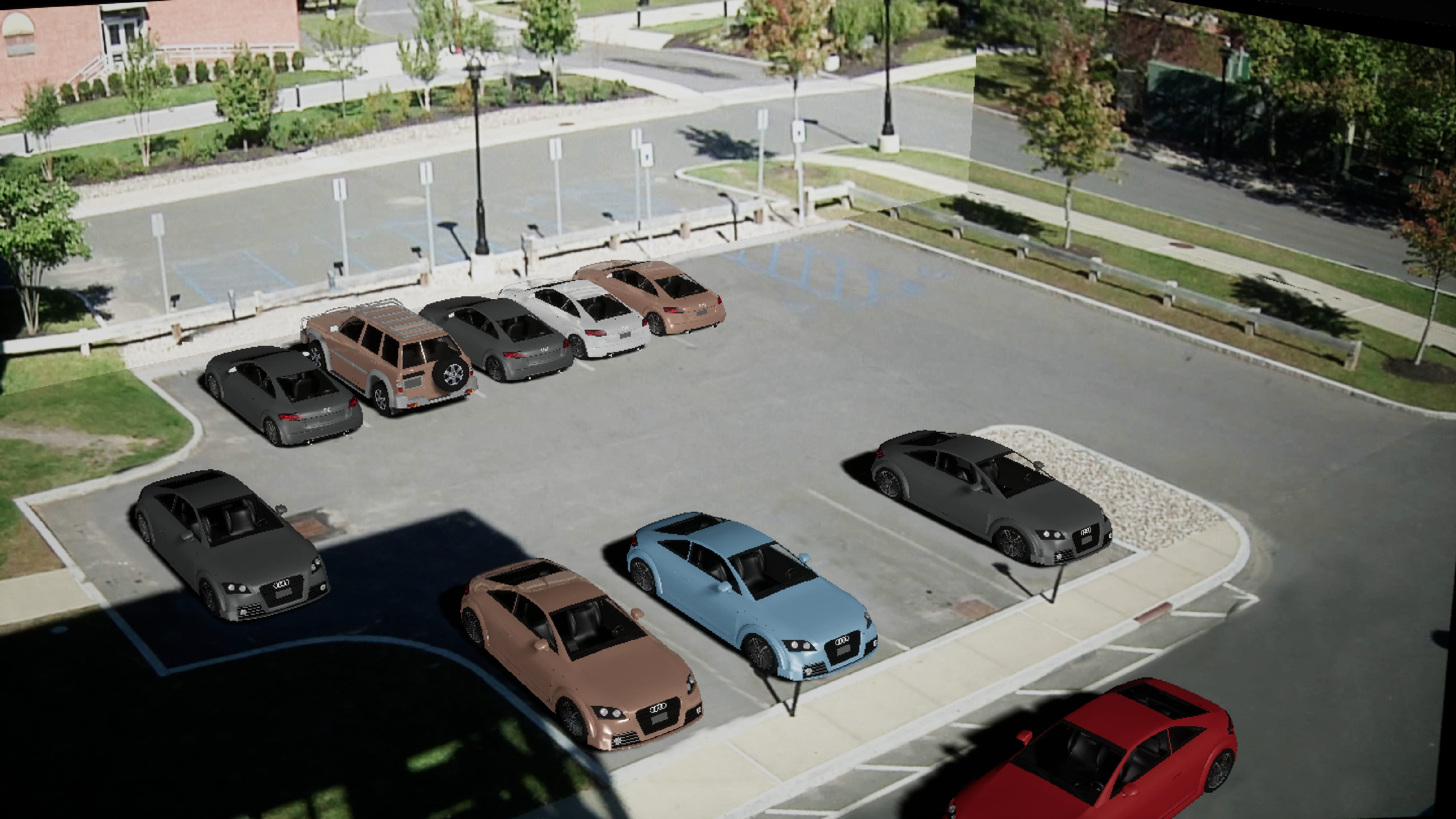}
\label{fig:dr_intro}
\\
\\
(b) Car detection and pose based 3D reconstruction
\\
 \end{tabular}
}
\end{center}
  \caption{
  Our results from a model trained only on synthetic data using domain randomization. (a) object detection results (b) 3D scene reconstruction using detection and pose estimations from our model.
  }
\label{fig:dr_intro}
\end{figure}




Consider the scenario in which a surveillance system is installed in a novel location and an image-based car detection and pose estimation model must be trained. A common approach is to frame this as a supervised learning problem and re-use models pre-trained on large visual datasets such as COCO \cite{lin2014coco}, KITTI \cite{geiger2013vision}, PASCAL3D+ \cite{xiang2014beyond}. However, the data distribution of these off-the-shelf datasets can vary significantly from the in-focus surveillance setting. For instance, the camera elevations of a close circuit camera are different from a vehicle mounted camera, a representative case of the KITTI dataset. Fine-tuning a pre-trained model using a small scene-specific dataset is often performed to ensure good performance during inference. We would like to highlight two major shortcomings of this approach, (1) obtaining a new dataset for each new surveillance location is not scalable due to expensive human hours spent in gathering these annotations, (2) another problem is overfitting to this small domain specific dataset during training. In this work, we provide an alternative solution to address these shortcomings using a synthetic data generation pipeline which is scalable and robust to data variance. Our experiments show that this annotation-free technique outperforms the baseline model trained using limited real data and achieves comparable performance to the model trained using real annotated data. 

Our goal is to develop a method for training a scene-specific car pose estimation and detection model without real annotations. This is an ill posed zero-instance learning problem, however, we do have access to three important priors here: (1) 3D scene geometry, (2) camera parameters and finally (3) priors on the geometry and appearance of the object of interest (such as cars) available in the form rich 3D CAD models. We model this domain specific knowledge in the form of a synthetic data generation pipeline to compensate for the absence of real training data. In this pipeline, we model the scene in 3D using state-of-the-art rendering engine Unreal Engine 4. An upside of using synthetic data is that we are capable of generating pixel-level accurate annotations like instance segmentation, depth map, category labels, 3D object locations and pose annotations. However, an underlying challenge here is that using synthetic data as the sole source of supervision can result into poor performance during inference on real data.

This is the well studied problem of 'domain shift' \cite{pan2010survey, wang2018deep, patel2015visual, csurka2017domain}. The model trained on synthetic data can over-fit to it, not generalizing well to real data. Majority of the approaches addressing 'domain shift' can be categorized into (1) domain adaptation and (2) domain randomization. Domain adaptation aims to learn features which helps generalization to target domain, thus modifying the learning model. On the other hand, underlying principal of domain randomization is to create enough variance in training data which forces the model to only learn relevant features useful for the task. Domain randomization provides a solution for domain shift in the data generation phase. Both these paradigms are proven to be successful by various works, however, in our work we prefer domain randomization over domain adaptation. This design decision allows us to modularize our solution nicely into two modules, data generation and learning model. Only the data generation module addresses domain shift, which provides us flexibility of making different design choices for our learning model. We provide an end-to-end framework capable of supporting multiple learning models which is an important attribute of a scalable system.

\textbf{Contributions:} The contributions of our work are as follows: 
\begin{itemize}
    \item Synthetic data generation pipeline: We encode known information like 3D scene geometry, camera parameters, object shape and appearance into rich annotated data.
    
    \item Domain randomization and a supportive ablation study: We use domain randomization to generate diverse synthetic data to span the space of realistic images. We conduct an ablation study to examine the effect of individual randomization components like texture randomization, light augmentation and distractors. 
    
    \item Evaluations on diverse datasets: We benchmark our approach on VIRAT~\cite{sangmin2011virat}, UA-DETRAC~\cite{lyu2017ua}, EPFL Cars Datasets~\cite{ozuysal2009epfl} for the task of car detection and pose estimation.
\end{itemize}

\section{Related Work}
Our work is related to object detection and pose estimation, synthetic data for computer vision, domain adaptation and domain randomization.  

\vspace{1mm}
\noindent
\textbf{3D Models for Object Detection and Pose Estimation}
Car detection and pose estimation is a well studied problem in the literature(see e.g., \cite{movshovitz20143d}, \cite{mousavian20173d}, \cite{savarese20073d}). The classical problem of 6 DoF pose estimation of an object instance from a single 2D image has been considered previously as a purely geometric problem known as the \textit{perspective n-point problem (PnP)}. Several closed form and iterative solutions assuming correspondences between 2D keypoints in the image and a 3D model of the object can be found in \cite{lepetit2009epnp}. 



\vspace{1mm}
\noindent \textbf{Synthetic Data for Computer Vision Tasks} 
There are many researches using synthetic data for computer vision tasks. Dhome \textit{et al.} used synthetic models to recognize objects from a single image~\cite{dhome1993pose}. For pedestrian detection, computer generated pedestrian images were used to train classifiers~\cite{hattori2018pedestrian}. 3D simulation has been used for multi-view car detection~\cite{pepik2012dpm}~\cite{yair2014cf}~\cite{hejrati2014synthesis}. Sun and Saenko~\cite{sun2014adaptation} trained 2D object detector with synthetic data generated by 3D simulation.   

\vspace{1mm}
\noindent \textbf{Domain Adaptation}
Ganin and Lempitsky proposed a domain adaptation method where the learned feature are invariant to the domain shift~\cite{ganin2015unsupervised}. \cite{zhang2015learning} suggested a multichannel autoencoder to reduce the domain gap  between real and synthetic data.
SimGAN~\cite{shrivastava2017simgan} used domain adaption for training eye gaze estimation systems on synthetic eye images. They solved the domain shift problem of synthetic images using a GAN based refiner that converts the synthetic images to the refined images. The refined images have similar noise distribution to real eye images.


\vspace{1mm}
\noindent \textbf{Domain Randomization}
\cite{sadeghi2016cad2rl} used domain randomization to fly a quadrotor through indoor environments. \cite{dosovitskiy2016learning} trained an agent to play Doom and generalize to unseen game levels. \cite{tobin2017domain}, \cite{james2017transferring}, \cite{pinto2017asymmetric}, \cite{peng2017sim} used domain randomization for grasping objects.  Tremblay~\textit{et al.} performed car detection with domain randomization~\cite{tremblay2018training}. \cite{sundermeyer2018implicit} proposed object orientation estimation for industrial part shapes that is solely trained on synthetic views rendered from a 3D model. They used domain randomization to reduce the gap between synthetic data and real data.


 \begin{figure*}[!t]
 \begin{center}
  \includegraphics[width=\linewidth]{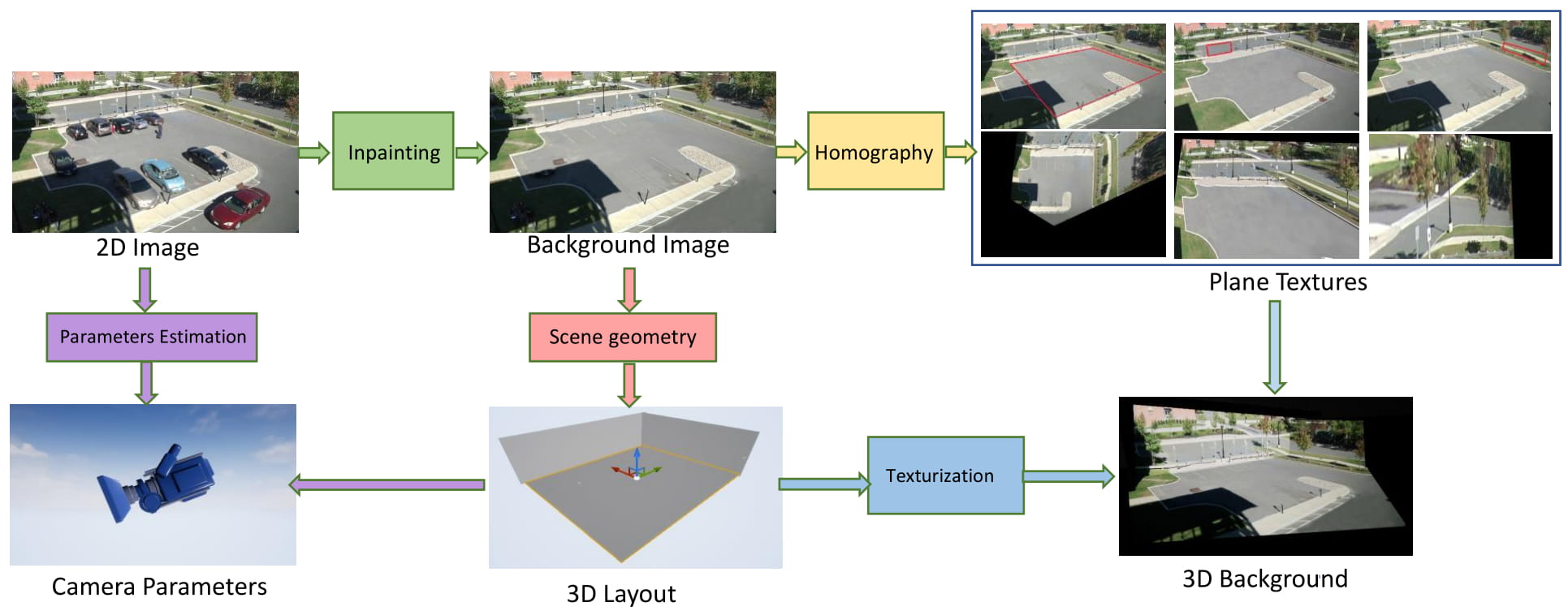}
 \end{center}
   \caption{3D background modeling from a single image with camera parameter estimation. Camera parameters are estimated and 3D layout is constructed from a 2D image using minimal user supervision. A background image without any objects is created using in-painting and plane textures are extracted using homography from this background image. Finally, we create the 3D background model by texturizing the layout with the extracted textures.
   }
\label{fig:pipeline1}
 \end{figure*}
 
 \begin{figure*}[!t]
 \begin{center}
  \includegraphics[width=\linewidth]{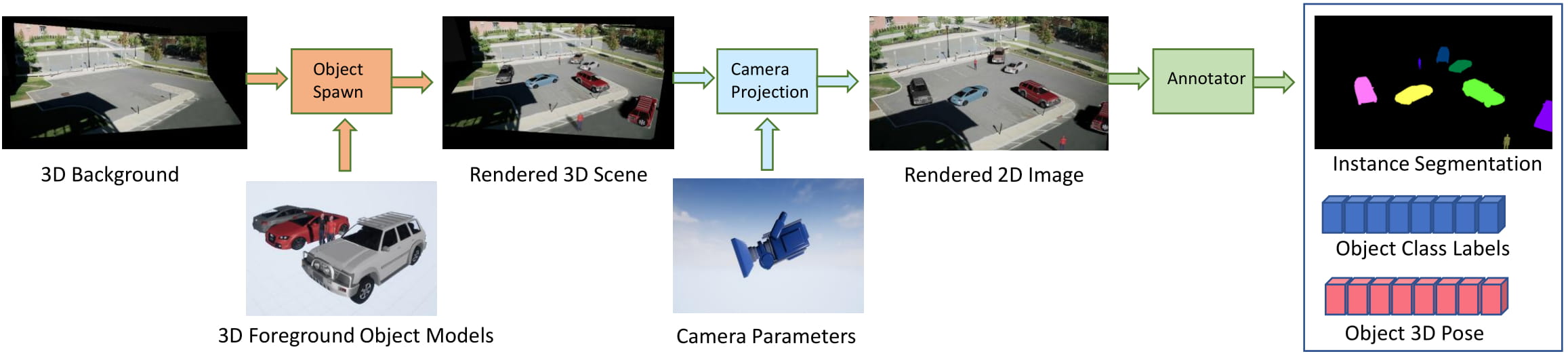}
 \end{center}
   \caption{Synthetic data and ground truth generation steps. Foreground object models are spawned in the 3D background model and 2D images are rendered according to the camera parameters from the previous step. The annotator module generates ground truths like bounding boxes, class labels, 3D poses, and instance segmentation using ray-tracing.}
 \label{fig:pipeline2}
 \end{figure*}
 
\section{Method Overview}
Our goal is to detect all the cars and estimate their orientations in a specific scene. We assume that scene geometry and camera parameters are given, and use that information along with a rich library of 3D CAD models to generate annotated synthetic data. As alluded to earlier, the main challenge which must be addressed is how to ensure generalization of our network trained on synthetic data to unseen real data. To address this challenge, we employ domain randomization in the data generation step to create synthetic data with enough variations such that the trained network does not over-fit to the synthetic data distribution, thus generalizing to real data during inference.

In this section, we first describe our scene-specific data generation methodology followed by details about domain randomization.

\begin{figure*}[t!]
\begin{center}
 \includegraphics[width=1\linewidth]{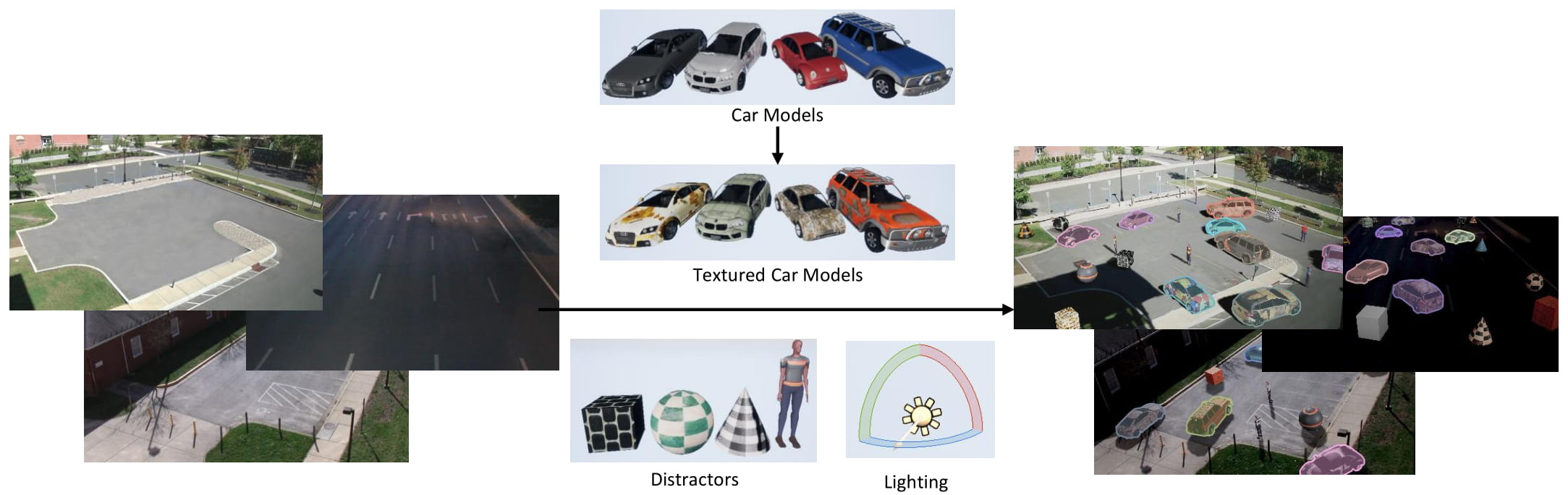}
\end{center}
   \caption{Annotated synthetic data generation with domain randomization: changing texture, lighting conditions, and addition of distractor objects. The instance segmentation map is overlayed over the RGB image}
\label{fig:dr}
\end{figure*}

\subsection{Synthetic Data Generation}
We use the state-of-the-art rendering platform Unreal Engine 4 for data generation. The first step is to encode the known scene geometry and camera parameters in our model with minimal user supervision. Figure \ref{fig:pipeline1} shows a schematic overview of this step. Here, we assume access to one image of the scene from the surveillance camera. First foreground objects are removed from this image using in-painting (or background extraction from video) giving us the scene's background image. A textureless 3D layout is constructed from the known scene geometry. The layout is then textured using homographic projections of scene's background image into different views giving us an approximate box-shaped 3D model of the background. The extrinsic camera parameters are estimated using 2D to 3D point correspondences provided by the user, we also perform grid search over an initial estimate of camera intrinsics if the knowledge of these exact parameters are not known. The scale transformation of an unit measure in the real scene to the synthetic scene is provided by the user. Finally, these assets, namely the camera and 3D textured layout represent the known camera parameters and scene geometry respectively.

We now proceed to encode rich priors about the foreground objects appearing in the scene using a large collection of 3D CAD models (see Figure \ref{fig:pipeline2}). 3D models in standard mesh representations are very high dimensional but interpretable and finite dimensional. By virtue of design, this gives us excellent control over the object's characteristic properties like scale, shape, texture. We now proceed to render foreground objects with desirable properties in the 3D layout. This flexibility is critical for our domain randomization setup. After the completion of rendering, we use Unreal Engine's ray tracing to generate pixel level accurate ground truth labels like instance segmentation, class labels and pose annotations.


\subsection{Domain Randomization}
The key idea here is to generate enough variations in the image space to avoid model overfitting. Also, the variations using domain randomization make the model robust to object appearances and light condition.

Concretely, we introduce randomization by varying the following aspects of the scene.

\vspace{1mm}

\begin{itemize}
    \item \textbf{Content Variation}: A random number of objects are placed randomly in the 3D background with varying orientation. To achieve shape and size variations we use an array of 5 different types of car models, the model's dimensions are randomly perturbed to provide more fine geometric variations.
    
    We also place 3D distractor objects in the scene like pedestrians, cones, spheres at varying dimensions to model the appearance of uninteresting objects in the scene.
  
    \item \textbf{Style Variation}: The style variation is achieved by randomly varying the color, texture of all the objects. We use a texture bank of 50 textures for this purpose.
    
    Furthermore, we apply light augmentations to capture the varying shadow conditions and time of day changes, we also varying the lightning conditions by changing the number of sources along with the location, orientation and luminosity. The generated image also undergoes random changes in contrast and brightness to model slight appearance changes of the background. 
\end{itemize}

\begin{figure*}[t]
\begin{center}
\begin{tabular}{ccc}
 \includegraphics[width=0.31\linewidth]{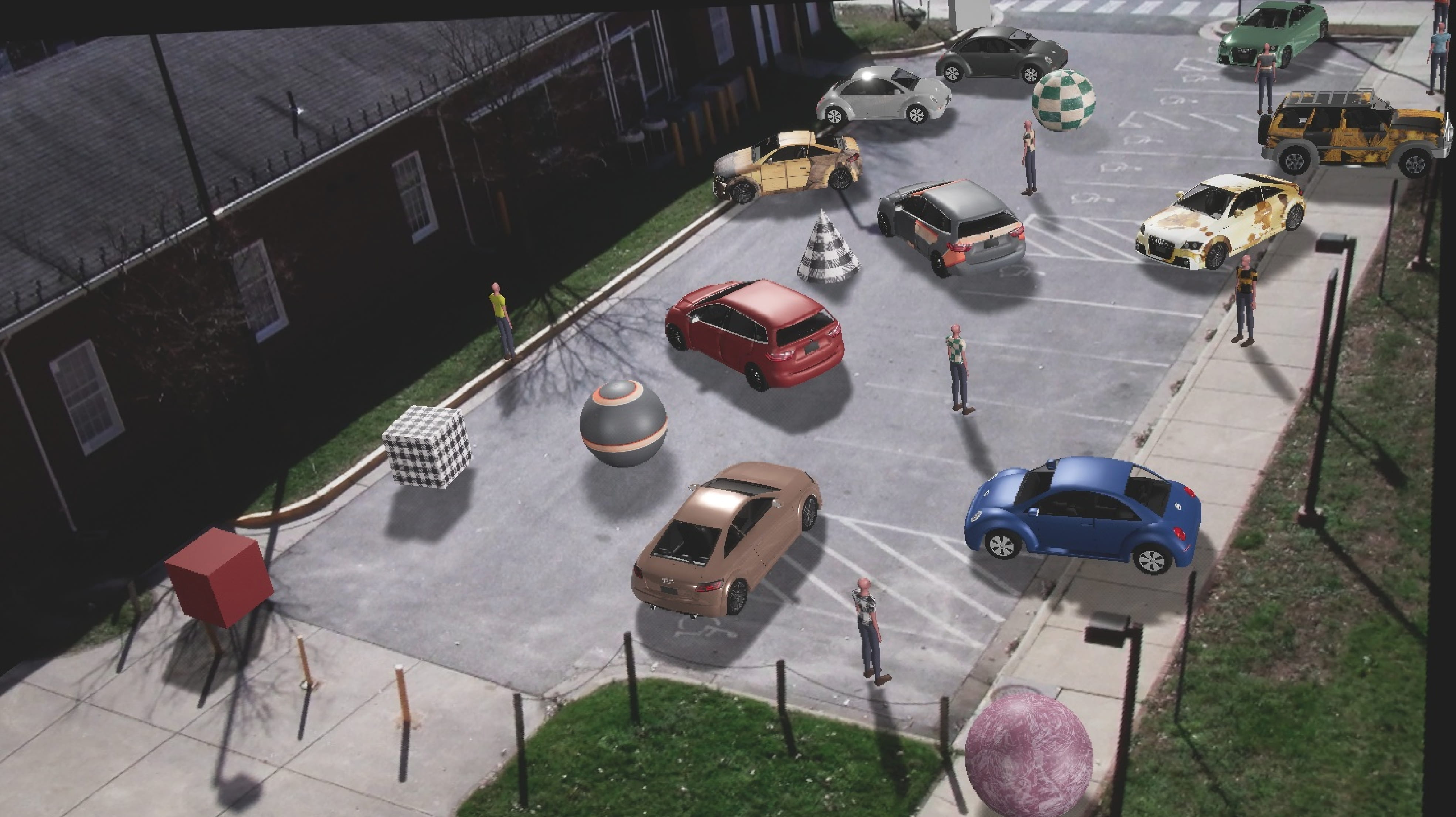} &
 \includegraphics[width=0.31\linewidth]{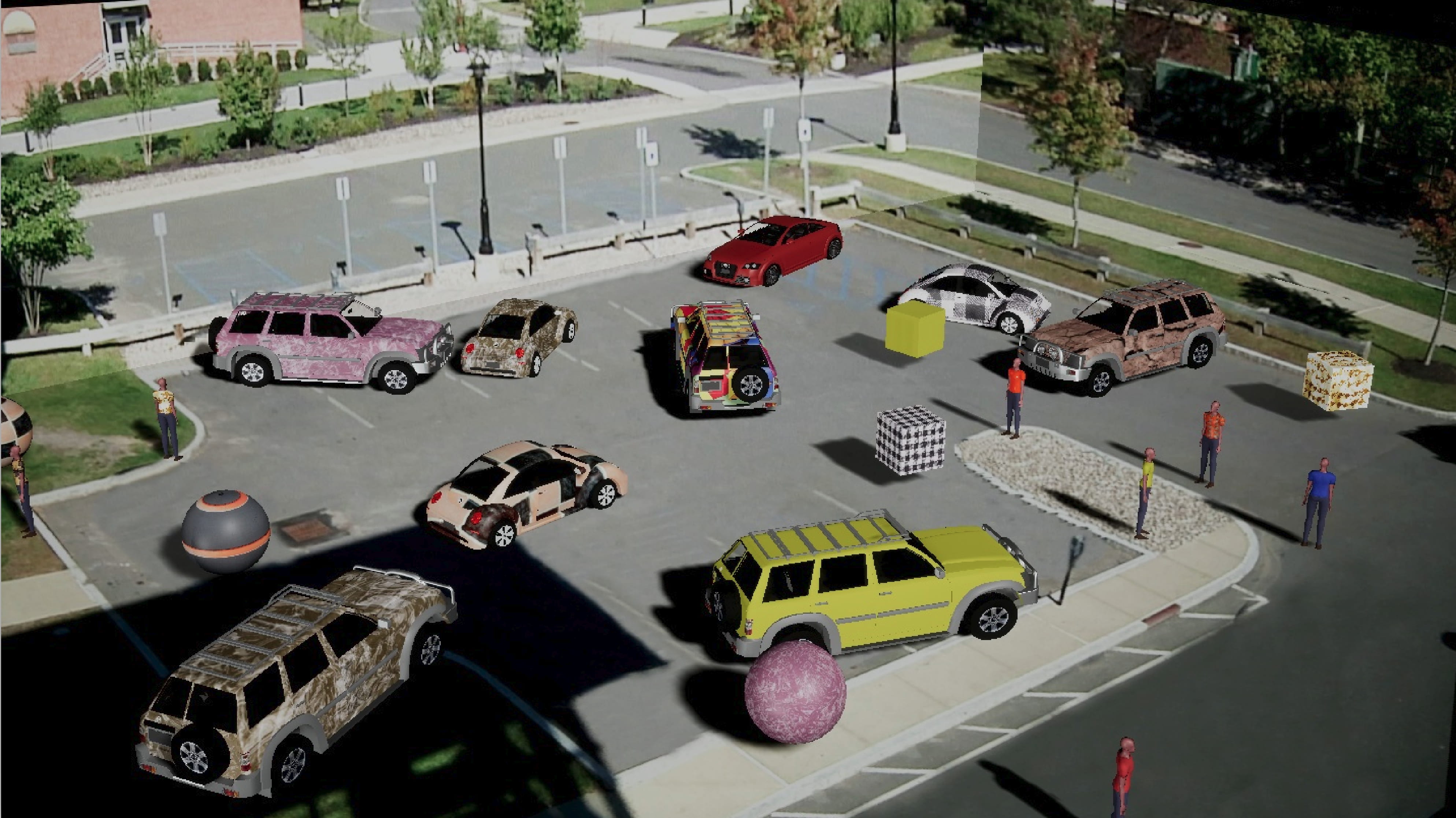} &
 \includegraphics[width=0.31\linewidth]{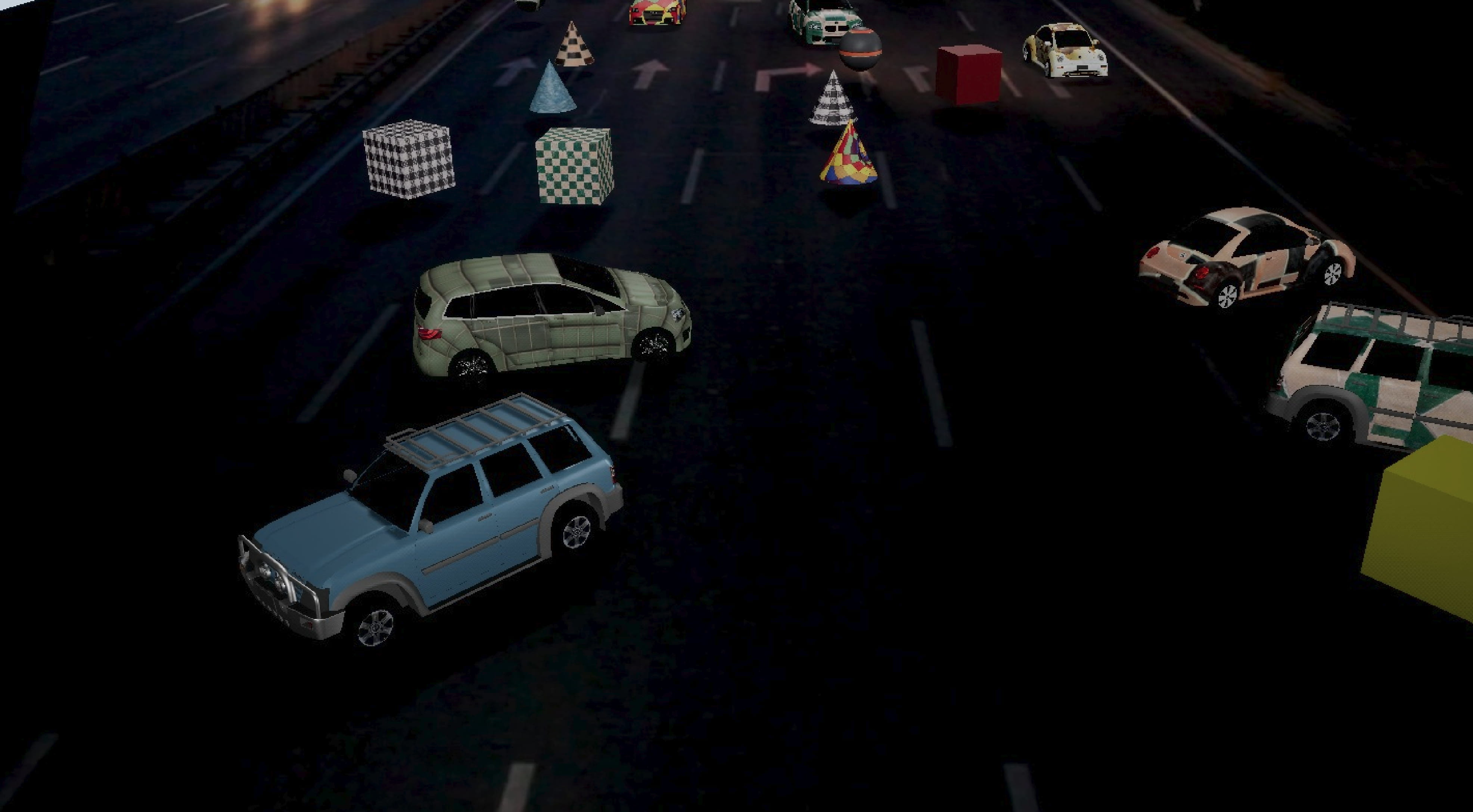}
 \end{tabular}
\end{center}
   \caption{Scenes rendered with domain randomization. The left and middle scenes are from VIRAT dataset and the right scene is from UA-DETRAC dataset. The foreground object textures and lighting conditions are randomized. Distractors such as cubes, spheres, cones and pedestrians are placed to model scene's variations.}
\label{fig:dr_samples}
\end{figure*}

\section{Detection and Pose Estimation Network}\label{model}

\begin{figure}[H]
\begin{center}
 \includegraphics[width=1\linewidth]{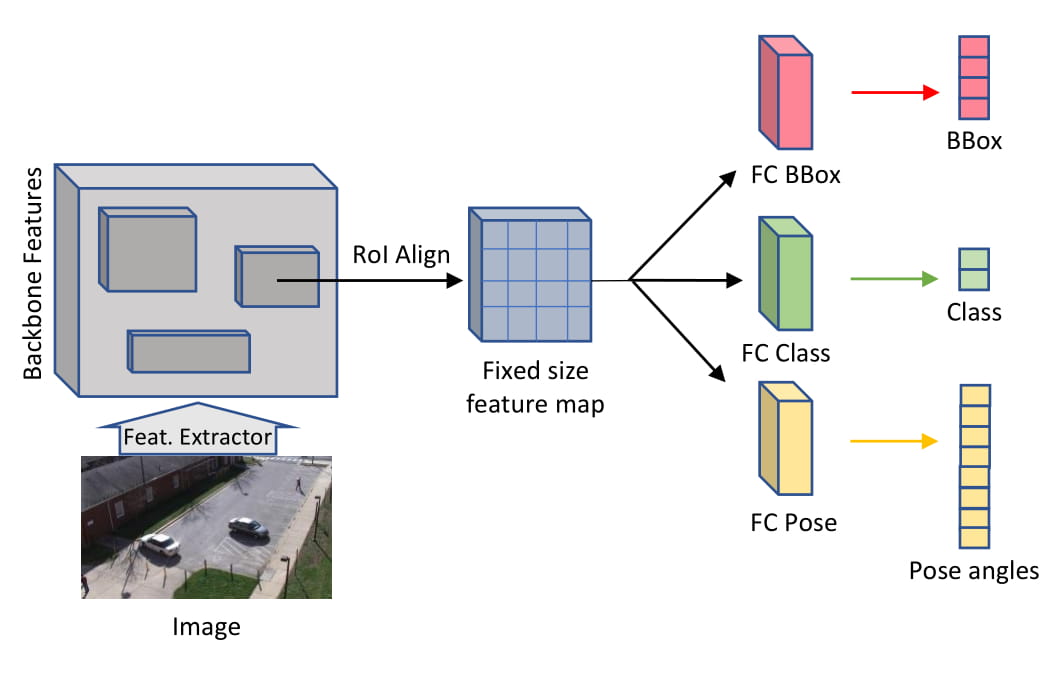}
\end{center}
   \caption{Car detection and pose estimation network that adopts Faster-RCNN architecture.}
\label{fig:model}
\end{figure}

For the learning model, our method adopts the Faster-RCNN~\cite{ren2015faster}/Network-on-Convolution meta-architecture. The network consists of a shared backbone feature extractor for the full-image, followed by region-wise sub-networks (heads) that predict the car's yaw angle with respect to the camera coordinate system in addition to traditional 2D bounding box and class label.

Our initial experiments indicated framing the problem of pose estimation as a classification task more favorably than as a regression task. Classification over-parametrizes the problem, and thus allows the network more flexibility to learn the task. As, the pose angles are bounded by $[-\pi,\pi]$, we quantize this space into 36 bins.

For our backbone network, we use the Feature Pyramid Network setup with ResNet101 architecture. Figure \ref{fig:model} illustrates our model. We define a multi-task loss function ($L$) for sampled region proposal as follows:

$L = L_\text{class} + L_\text{bbox} + L_\text{pose} $

\noindent
where $L_\text{class}$ is the object classification loss, $L_\text{bbox}$ is the bounding box regression loss as defined in \cite{ren2015faster}. $L_\text{pose}$ is the pose label classification loss.

\section{Experiments}
We benchmark our methodology on three datasets VIRAT, EPFL-Car and UA-DETRAC for the tasks of car detection and pose estimation.  

\begin{itemize}
    \item VIRAT dataset is designed to be realistic, natural and challenging for video surveillance domains in terms of its background clutter, diversity in scenes. The captured video are recorded at 25 frames per seconds (fps) at the resolution of $1920 \times 1080$ pixels. 
    \item EPFL-Car dataset contains 20 sequences of cars as they rotate by 360 degrees. The dataset contains images with variation in background, and slight changes in camera elevation recorded at $376 \times 250$ pixels.  
    \item UA-DETRAC is a challenging real-world multi-object detection benchmark consisting of 10 hours of videos captured at 24 different locations. The videos are recorded at 25 frames per seconds (fps), with resolution of $960 \times 540$ pixels.
\end{itemize}

We perform evaluations on two scenes from VIRAT dataset, entire EPFL-Car dataset and two scenes from UA-DETRAC dataset. 10,000 synthetic images are rendered for each scene at the resolution of $1440 \times 810$ pixels for each variation in the ablation study with 80:20 split as the training and validation set. Furthermore, 8000 images per scene (VIRAT), 2300 images (EPFL-Car), 2000 images per scene (UA-DETRAC) constitutes our real data for evaluations. A random 70:20:10 split of the real images per scene makes up our training, validation and test set respectively.  

Note that to capture the background and camera elevation changes in the EPFL-Car dataset, we also randomly perturb the camera parameters and background textures during data generation.

For all our experiments, we compare six models. 
\begin{itemize}
\itemsep0em
    \item SS-scratch, initialized with random weights and trained on scene specific synthetic data without domain randomization.
    \item SS-finetune, initialized with COCO and fine-tuned on scene specific synthetic data without domain randomization.
    \item SS+DR-scratch, initialized with random weights and trained on scene specific domain randomized data.
    \item SS+DR-finetune, initialized with COCO and fine-tuned on scene specific domain randomized data.
    \item Real-0.5, initialized with COCO and fine-tuned on 50\% real training data.
    \item Real, initialized with COCO and fine-tuned on the entire real training data.
\end{itemize}

Without domain randomization: We use 4 car models with fixed dimensions, 7 colors, constant lighting conditions without contrast or brightness augmentation. We do not use cubes, spheres or cones as distractors, however pedestrians are still rendered.

To ensure consistency, we use similar model architecture as reported in section \ref{model}, same hyper-parameters and training regime. We also compare our models against the baseline of an off-the-shelf pre-trained network for the car detection and pose estimation.

\begin{figure}
\begin{center}
 \includegraphics[width=1\linewidth]{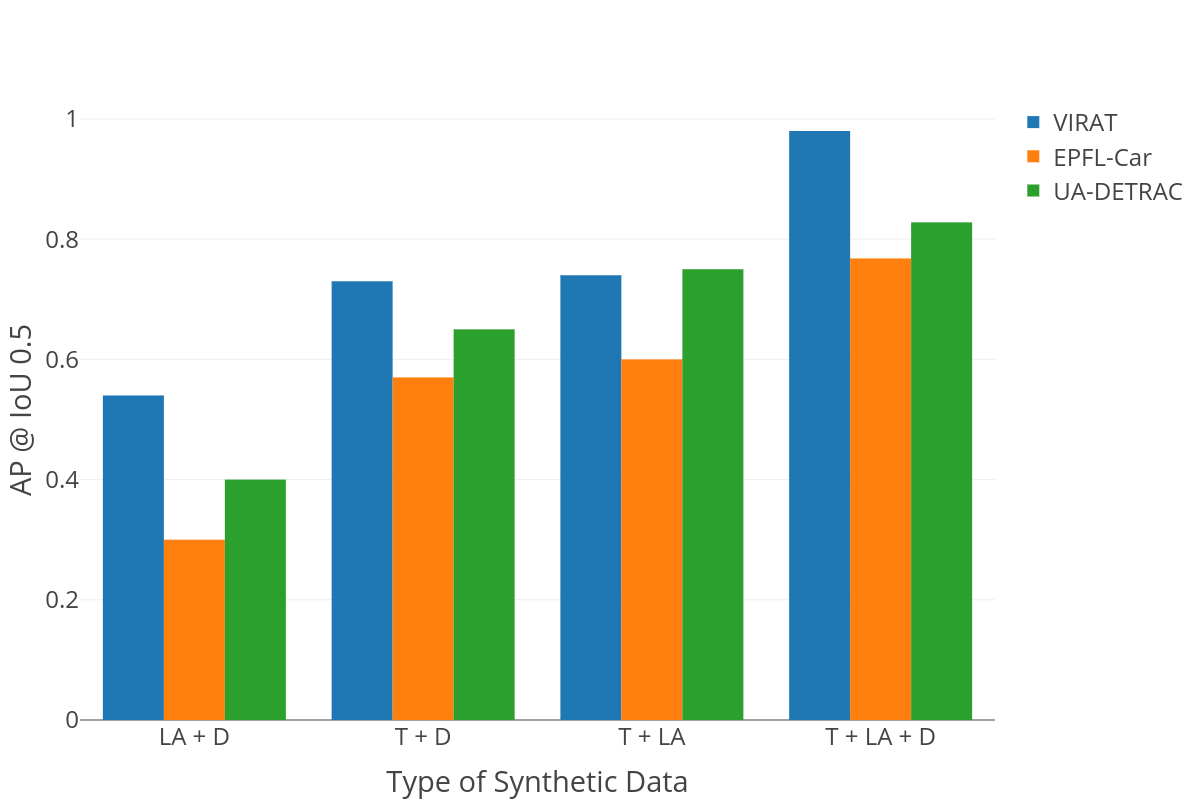}
\end{center}
   \caption{Detection ablation study across three datasets. Y axis is AP\@0.5, X axis various randomization conditions} 
\label{fig:bb_ablation}
\end{figure}

\subsection{Object Detection}
The pre-trained model we use in detection evaluations is a car detector trained on 5010 car images from the COCO dataset (referred as COCO henceforth). We report average precision for bounding box regression at IoU=\{0.5, 0.75\}.

\vspace{2mm}
\noindent
\textbf{Quantitative Analysis}:
Table \ref{tab:bb_table} shows the detection results tested on VIRAT, EPFL-Car, and UA-DETRAC dataset. Our SS+DR-finetune achieves the best performance in the most cases and outperforms Real-0.5. We hypothesize this is due to overfitting on limited real data which is avoided by design in domain randomization. Furthermore, our model trained only on synthetic data is competitive to the model trained on the entire real data at IoU=0.5. 

SS+DR-finetune outperforms the off-shelf baseline COCO by more than 6\% and this margin only increases when the scene distribution differs considerably from the off-shelf dataset's distribution. 

\begin{table*}[t]
\begin{center}
\resizebox{1\linewidth}{!}{
\begin{tabular}{|l||c|c||c|c||c|c||c|c||c|c||}
\hline
\multirow{2}{*}{Model}  & \multicolumn{2}{c||}{VIRAT Parking 1} & \multicolumn{2}{c||}{VIRAT Parking 2} & \multicolumn{2}{c||}{EPFL-Car} & \multicolumn{2}{c||}{UA-DETRAC Night} & \multicolumn{2}{c||}{UA-DETRAC Street}\\
 & AP@0.5 & AP@0.75 & AP@0.5 & AP@0.75 & AP@0.5 & AP@0.75 & AP@0.5 & AP@0.75 & AP@0.5 & AP@0.75\\
\hline\hline
COCO            & 
92.9            & \textbf{75.6}     & 85.5          & 43.0          & 
55.7            & \textbf{28.5}     &
62.9            & 34.5              & 78.3          & \textbf{51.8} \\

SS-scratch      &
54.2            & 37.8              & 49.1          & 18.9          &
31.5            & 0.4               &
51.0            & 20.4              & 36.7          & 12.8          \\

SS-finetune     &
80.3            & 39.5              & 54.6          & 19.2          &
38.1            & 3.5               &
52.5            & 21.9              & 38.3          & 15.0          \\

SS+DR-scratch      &
98.8            & 56.4              & 81.7          & 34.3          &
48.6            & 7.4               &
78.7            & 40.6              & 52.1          & 29.6          \\

SS+DR-finetune     &
\textbf{98.9}   & 62.6              & \textbf{97.6} & \textbf{48.8} &
\textbf{76.8}   & 22.9              &
\textbf{82.8}   & \textbf{56.7}     & \textbf{87.3} & 51.2          \\

\hline

Real-0.5        &
94.3            & 82.1              & 87.5          & 57.4          &
62.3            & 48.1              &
82.3            & 67.1              & 84.0          & 44.2          \\

Real            &
98.9            & 84.8              & 98.1          & 61.8          &
82.7            & 60.8              &
90.1            & 73.3              & 92.0          & 65.1          \\

\hline
\end{tabular}
}
\end{center}
\caption{Detection results tested on VIRAT, EPFL-Car, UA-DETRAC dataset. }
\label{tab:bb_table}
\end{table*}

\begin{figure*}[!t]
\begin{center}
\begin{tabular}{ccc}
COCO     & SS-finetune          & SS+DR-finetune \\
 \includegraphics[width=0.31\linewidth]{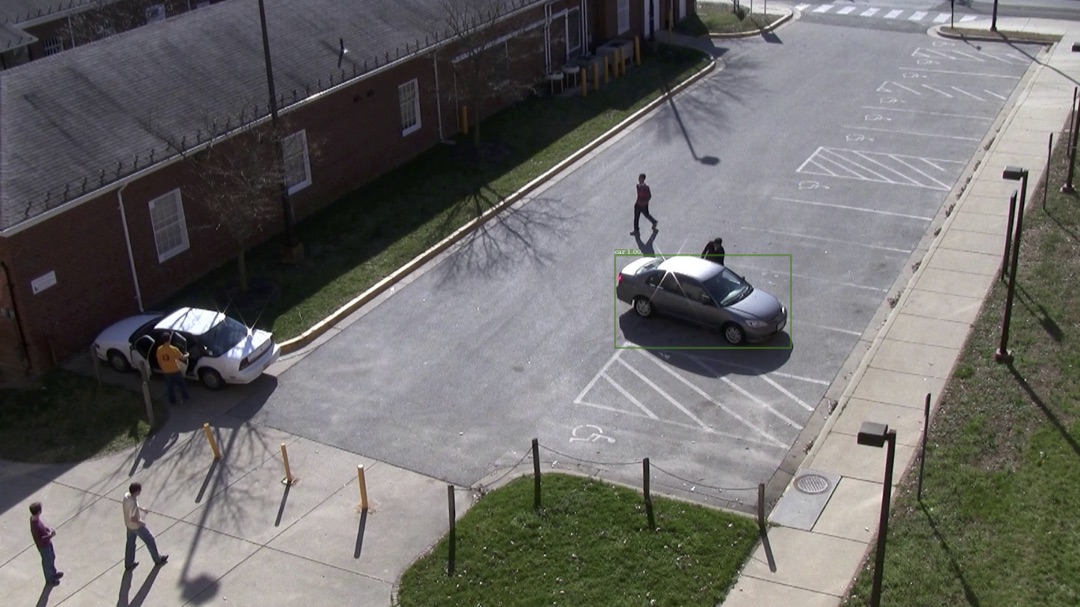} &
 \includegraphics[width=0.31\linewidth]{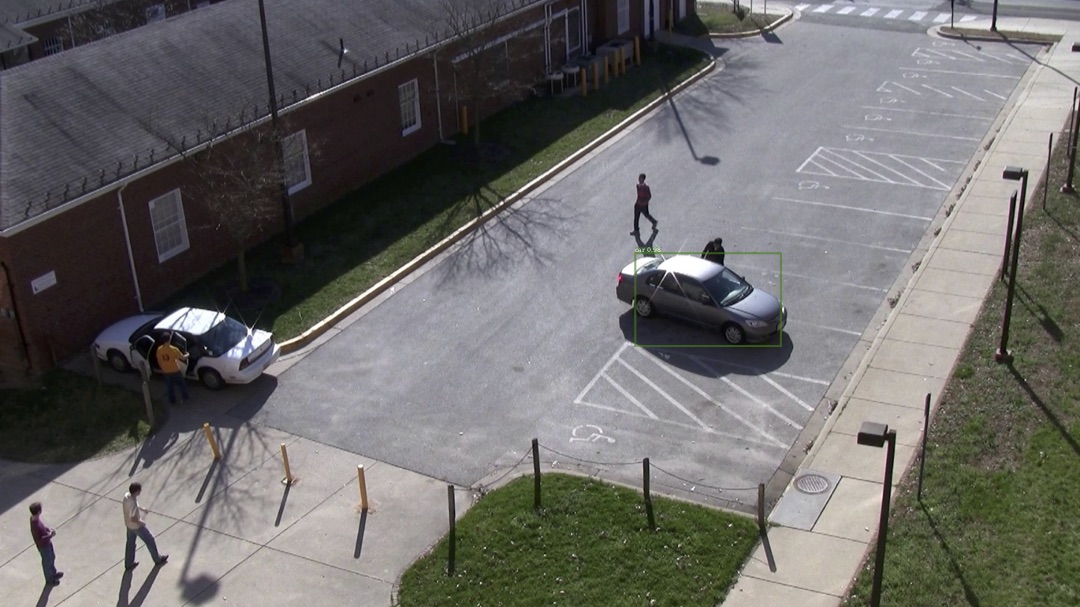} &
 \includegraphics[width=0.31\linewidth]{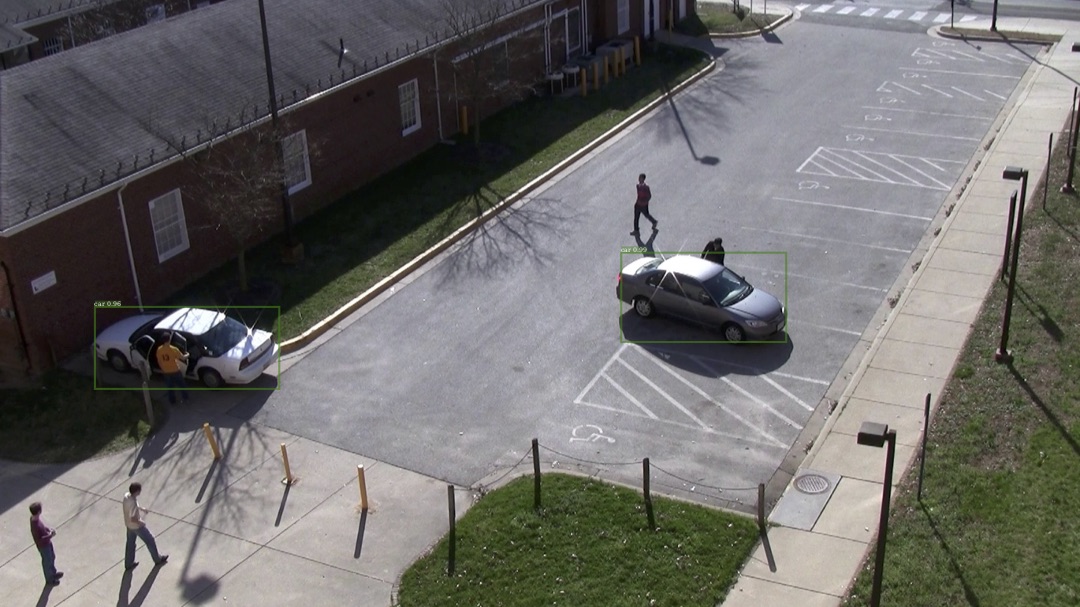}\\
 \includegraphics[width=0.31\linewidth]{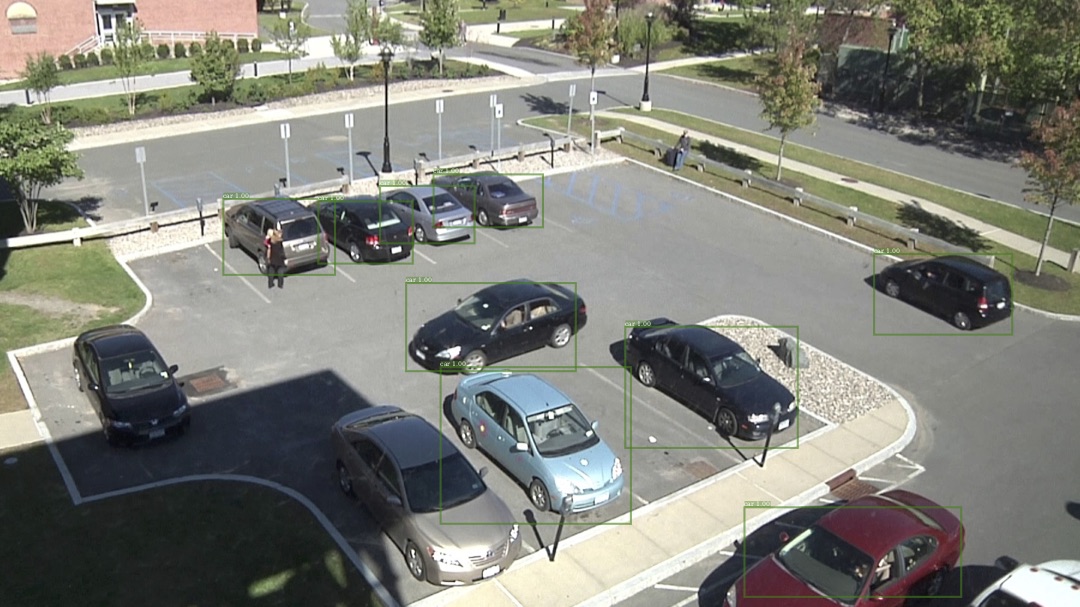} &
 \includegraphics[width=0.31\linewidth]{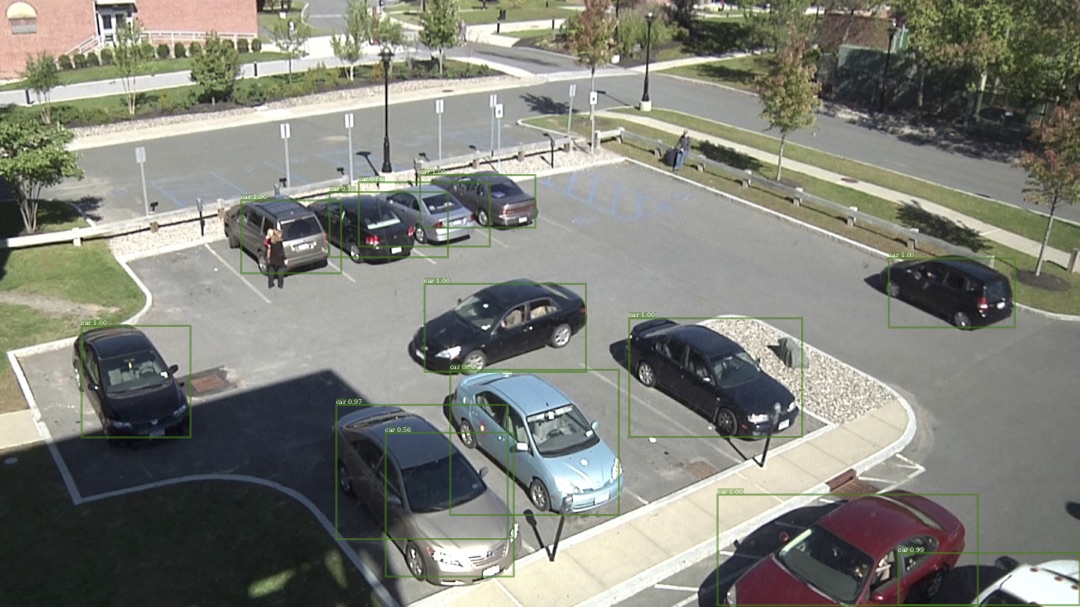} &
 \includegraphics[width=0.31\linewidth]{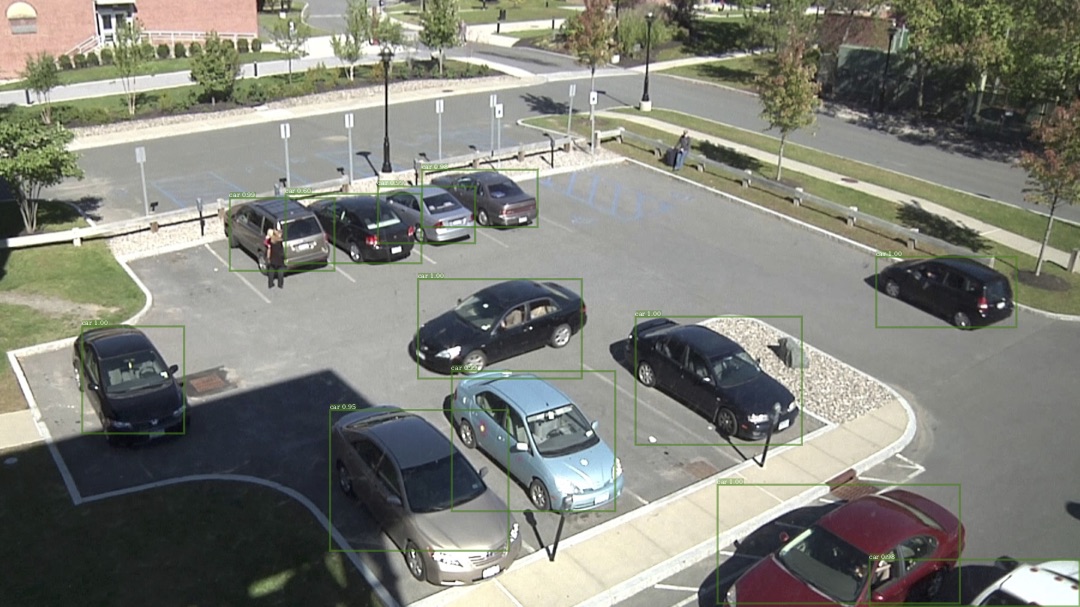}\\
 \includegraphics[width=0.31\linewidth]{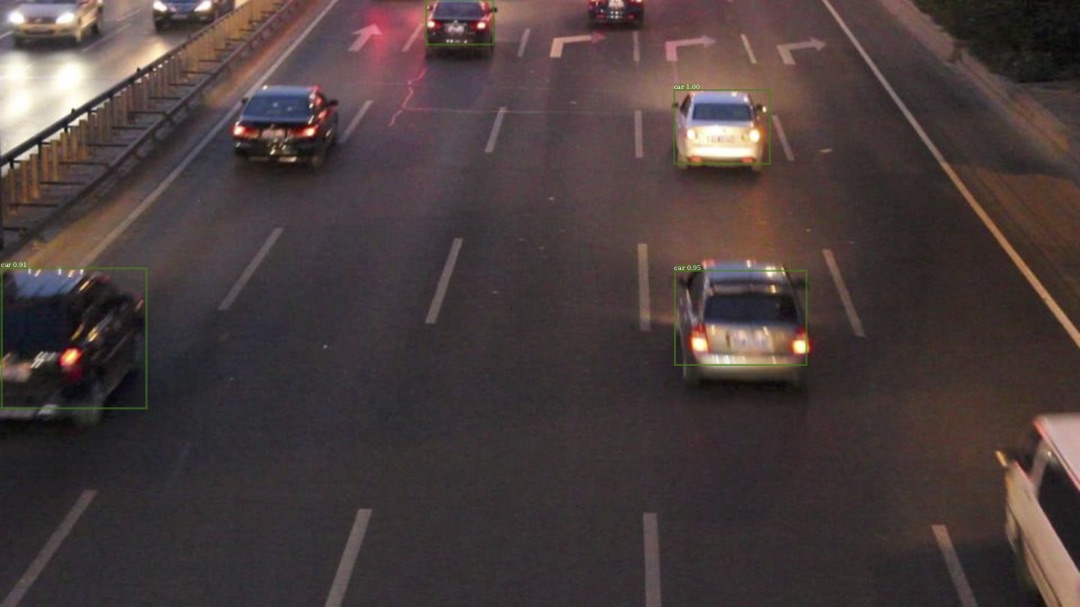} &
 \includegraphics[width=0.31\linewidth]{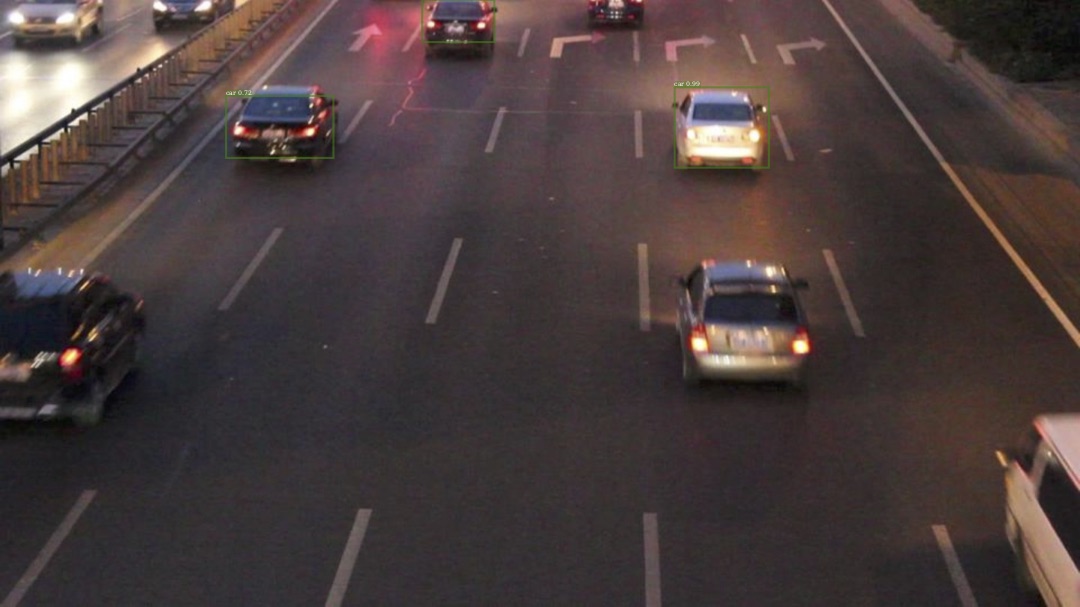} &
 \includegraphics[width=0.31\linewidth]{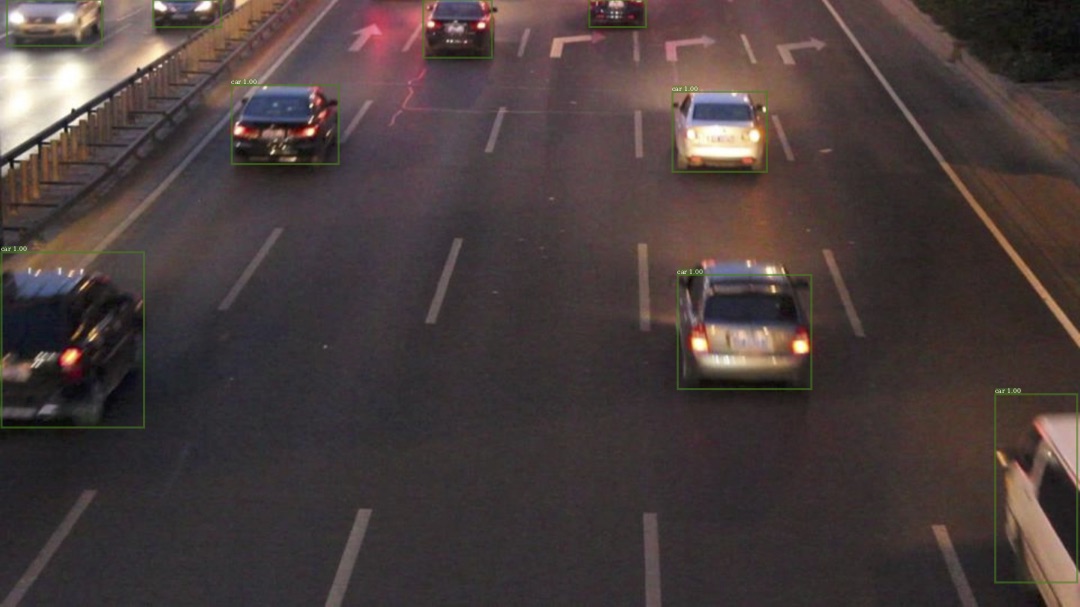}\\
\end{tabular}
\end{center}
   \caption{Examples of detection results. We compare the performance of COCO (first column) and SS-finetune (second column) with SS+DR-finetune (third column) on VIRAT Parking 1(top row), VIRAT Parking 2(second row), and UA-DETRAC Night (third row).}
\label{fig:qual_detection}
\end{figure*}

\vspace{2mm}
\noindent
\textbf{Qualitative Analysis}: Figure~\ref{fig:qual_detection} shows qualitative detection results, in this section we compare COCO, SS-finetune, and SS+DR-finetune. COCO fails in cases of severe object occlusion (first row), unusual shadows (first row, second row) and object truncation (second row, third row). Even though we explicitly model object occlusions and truncation in our synthetic data, SS-finetune fails to generalize to real data. This is because the deep network latches on to the synthetic dataset's biases, our SS+DR-finetune does not suffer from any of these shortcomings.

Furthermore, we found that the bounding boxes of SS+DR-finetune are more accurate than SS-finetune. We argue that this is because SS-finetune overfits to the height-to-width distribution of bounding boxes in the synthetic dataset. Domain randomization helps mitigates this problem by randomizing the dimensions and shapes of the 3D models during data generation, thus directly providing enough variance in the bounding box distribution. This regularizes SS+DR-finetune.  

\subsection{Pose Estimation}
Similar to COCO for detections, we benchmark our models against Deep3DBox \cite{mousavian20173d} trained on KITTI dataset for 3D bounding box estimation. To ensure consistency, we use ground truth detections as our region proposals for all our experiments and extract features using this proposals for the task of pose estimation.

\vspace{2mm}
\noindent
\textbf{Quantitative Analysis}:
Table \ref{tab:pose_table} shows the pose estimation results, we benchmark our results on EPFL-Car and UA-DETRAC datasets. EPFL-Car already has car pose annotations, we manually annotate two scenes from UA-DETRAC with pose annotations. We report pose classification accuracy $\text{Acc}_{10^o}\uparrow$ at $10^o$ degree quantization and median error angle MedErr$\downarrow$.

Our SS+DR-finetune achieves the best results that are competitive to Real-0.5 and Real in all the cases even though we don't use any real images for training. 


\begin{table*}[t]
\begin{center}
\resizebox{0.7\linewidth}{!}{
\begin{tabular}{|l||c|c||c|c||c|c||}
\hline
\multirow{2}{*}{Model}  &  \multicolumn{2}{c||}{EPFL-Car} & \multicolumn{2}{c||}{UA-DETRAC Night} & \multicolumn{2}{c||}{UA-DETRAC Street}\\
 & $\text{Acc}_{10^o}\uparrow$  & MedErr$\downarrow$ & $\text{Acc}_{10^o}\uparrow$  & MedErr$\downarrow$ & $\text{Acc}_{10^o}\uparrow$  & MedErr$\downarrow$\\
\hline\hline
Deep3DBox-KITTI & 
0.67            &$18.3^o$           &
0.41            &$28.0^o$           &0.58           &$17.7^o$       \\


SS-scratch      &
0.35            &$46.2^o$           &
0.18            &$63.1^o$           &0.45           &$38.3^o$       \\

SS-finetune     &
0.66            &$16.1^o$           &
0.51            &$31.8^o$           &0.72           &$23.0^o$       \\

SS+DR-scratch   &
0.72            &$10^o$             &
0.74            &$21.6^o$           &0.61           &$\mathbf{10.4^o}$       \\
SS+DR-finetune  &
\textbf{0.86}            &$\mathbf{6.6^o}$            &
\textbf{0.83}            &$\mathbf{14.2^o}$           &\textbf{0.87}           &$13.2^o$       \\

\hline

Real-0.5        &
0.91            &$5.4^o$            &
0.93            &$11^o$             &0.96           &$5.8^o$       \\

Real            &
0.96            &$6.1^o$            &
0.97            &$10.4^o$           &0.97           &$3.4^o$       \\

\hline
\end{tabular}
}
\end{center}
\caption{Pose estimation results tested on EPFL-Car and UA-DETRAC datasets. $\text{Acc}_{10^o}$ is pose label classification accuracy at 10 degree quantization and MedErr is the median error angle }
\label{tab:pose_table}
\end{table*}

\begin{figure*}
\begin{center}
\begin{tabular}{c|ccc}
Real Image & Deep3DBox-KITTI     & SS-finetune          & SS+DR-finetune \\
 \includegraphics[width=0.21\linewidth]{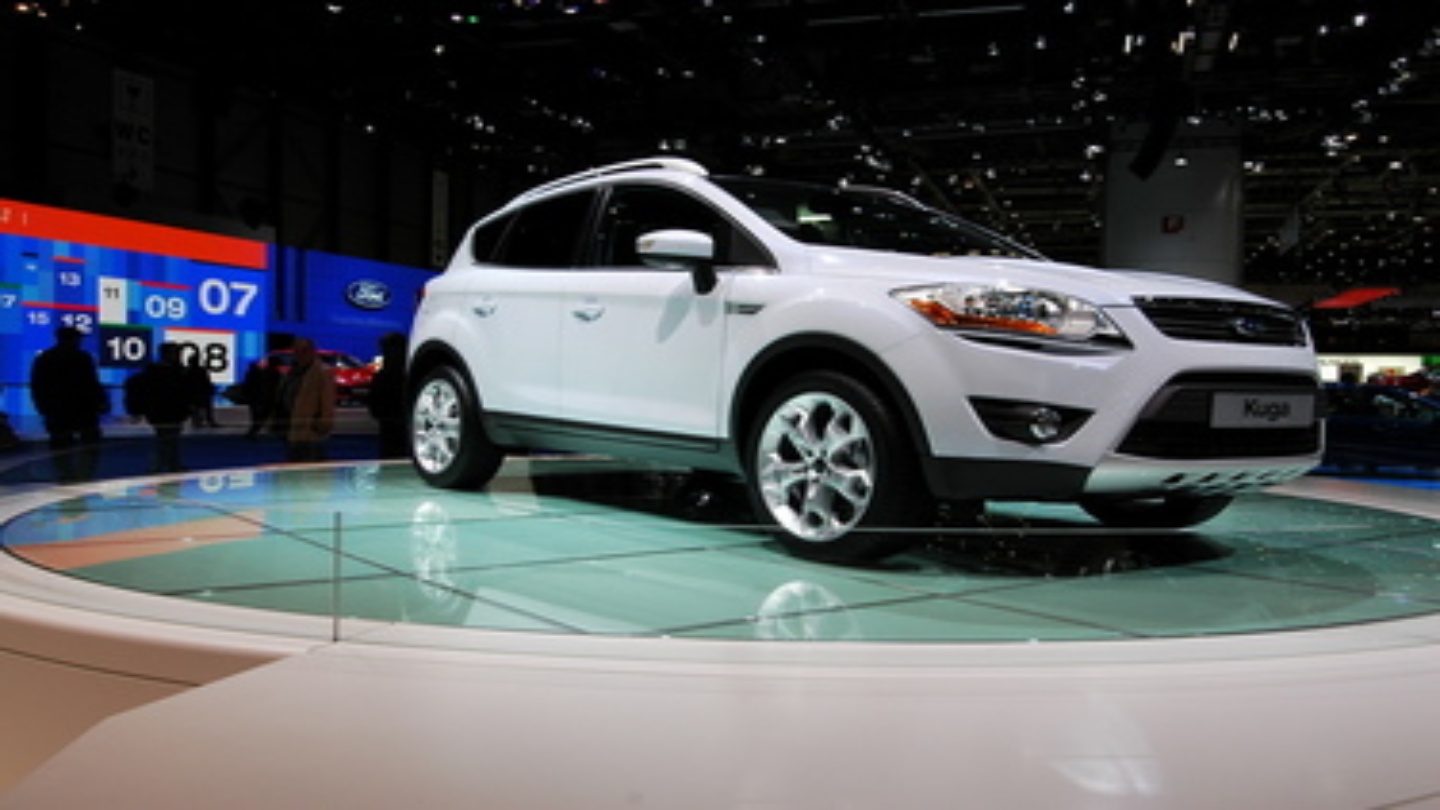} &
 \includegraphics[width=0.24\linewidth]{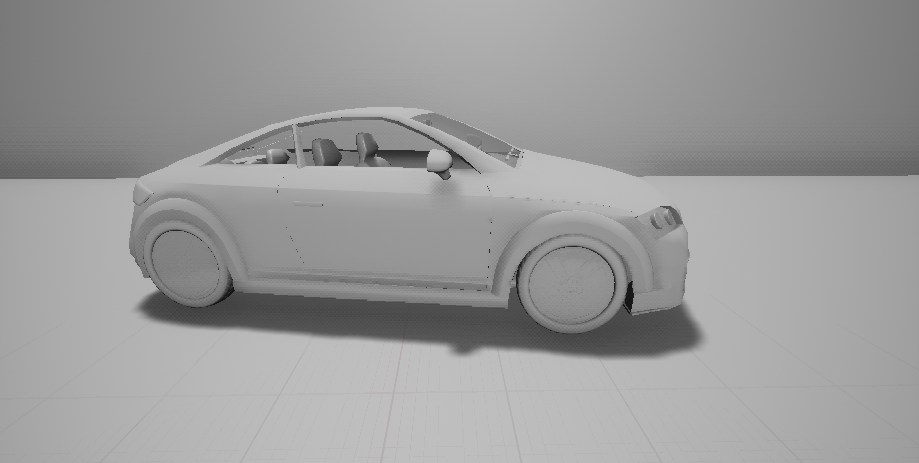} &
 \includegraphics[width=0.24\linewidth]{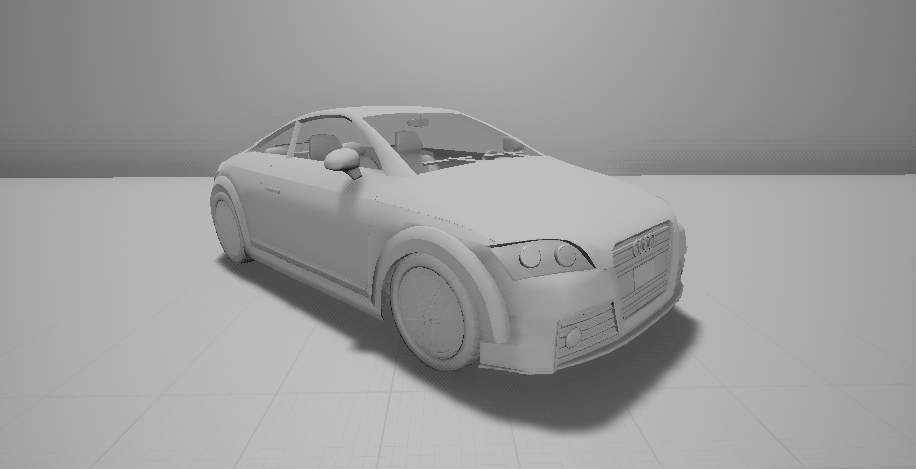} &
 \includegraphics[width=0.24\linewidth]{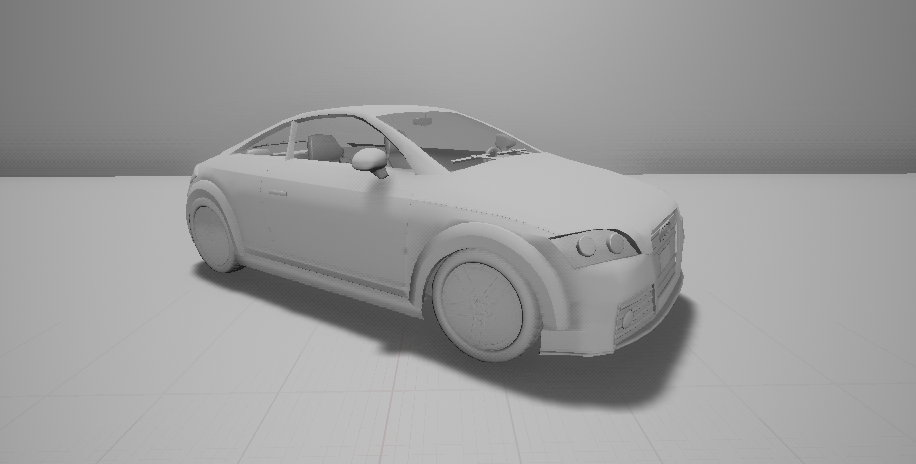}\\
 $\theta=308.76^o$ &
 $\theta=280.3^o$ &
 $\theta=315.0^o$ &
 $\theta=305.0^o$\\
 \includegraphics[width=0.23\linewidth]{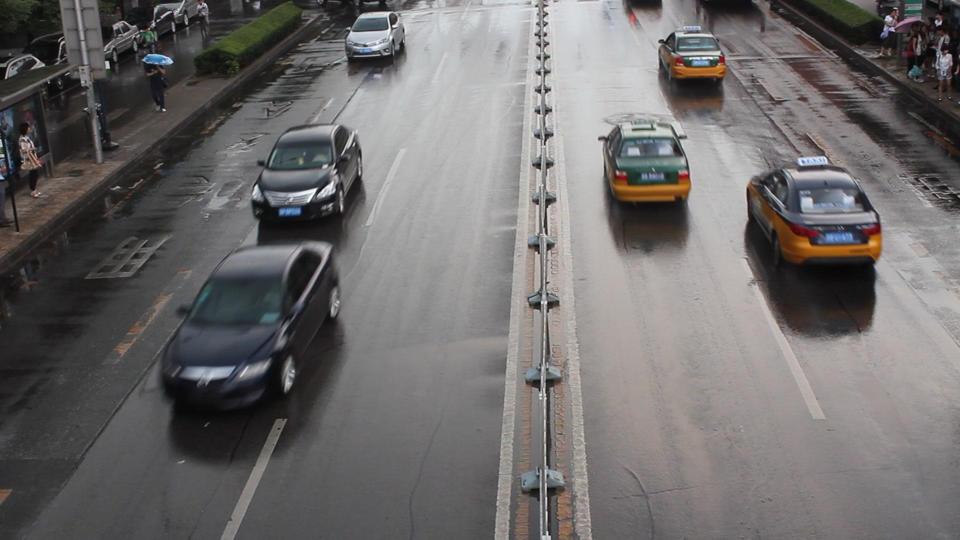} &
 \includegraphics[width=0.23\linewidth]{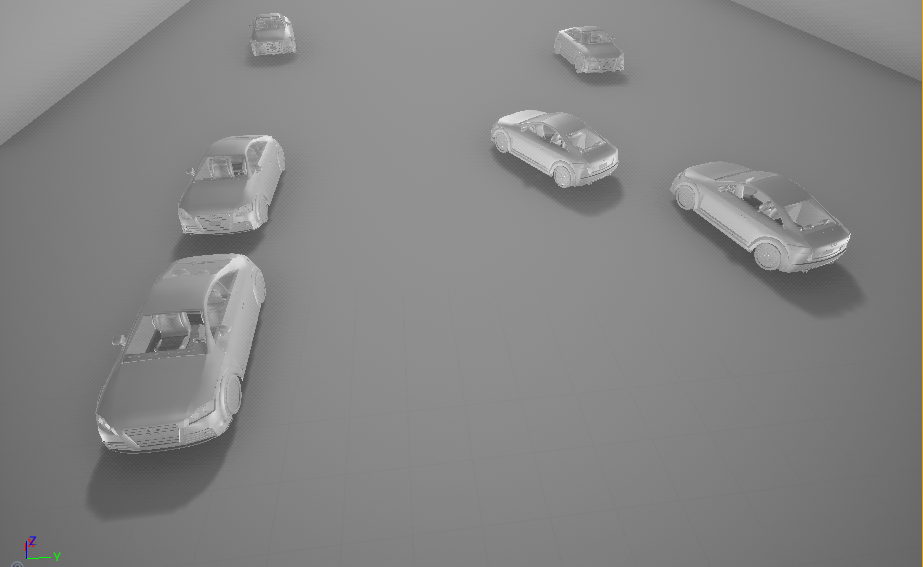} &
 \includegraphics[width=0.23\linewidth]{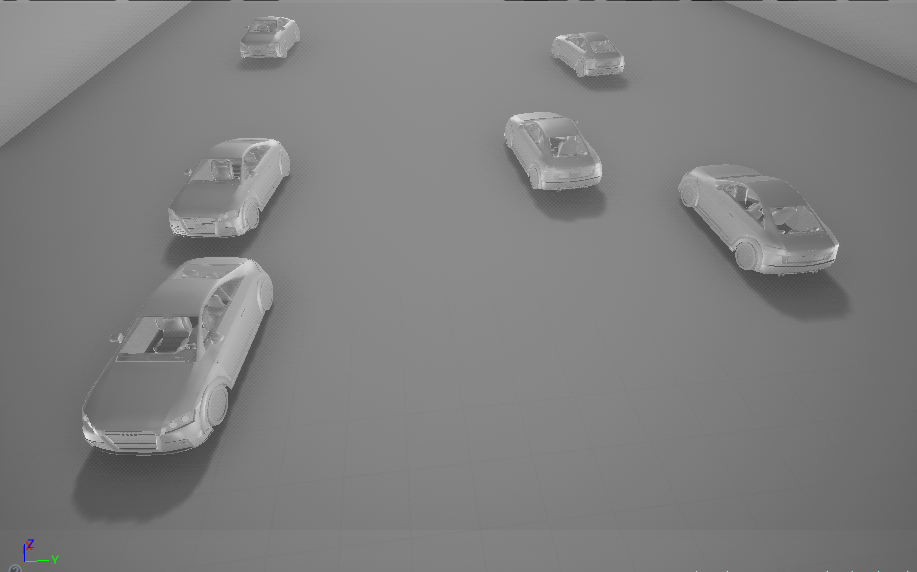} &
 \includegraphics[width=0.23\linewidth]{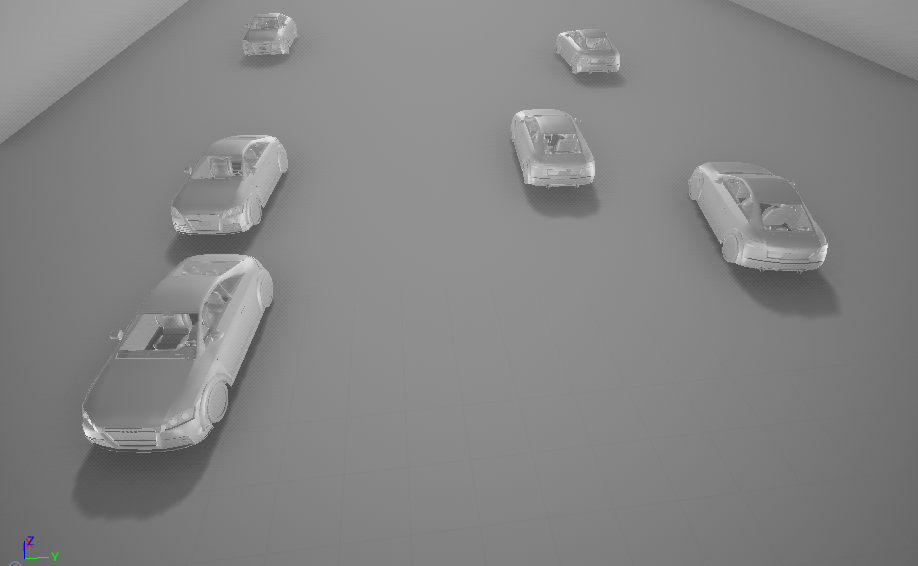}\\
 Mean Angle Error = $\bar{\theta}_{error}$&
 $\bar{\theta}_{error}=73.8^o$ &
 $\bar{\theta}_{error}=21.4^o$ &
 $\bar{\theta}_{error}=9.5^o$\\

\end{tabular}
\end{center}
   \caption{We compare the performance of KITTI pretrained model with our models on EPFL-Car(top row) and UA-DETRAC(second row). (1) Due to camera perspective difference between KITTI and EPFL-Car dataset, the pre-trained baseline performs poorly. (2) UA-DETRAC camera perspective on the other hand is much similar to KITTI, however the cars (yellow) in the UA-DETRAC dataset do not appear in the KITTI dataset. This illustrates overfitting to characteristics of different domain, the top right car is therefore flipped by ${180}^o$. }
\label{fig:pose_qualitative}
\end{figure*}

\vspace{2mm}
\noindent
\textbf{Qualitative Analysis}: 
Figure~\ref{fig:pose_qualitative} shows qualitative pose results, in this section we compare Deep3DBox-KITTI, SS-finetune, and SS+DR-finetune on images from EPFL-Car and UA-DETRAC Street scene. Note we assume ground truth bounding boxes as input for this analysis.

The camera perspective is significantly different for KITTI and EPFL-Car dataset, this results in poor performance of Deep3DBox on EPFL-Car images. However, SS-finetune performs better than Deep3DBox in our analysis, as we assume ground truth bounding box proposals, this is due to the task specific network learning appearance invariant features for pose prediction.

On the UA-DETRAC dataset, the camera perspective is much more similar to KITTI dataset, however the car appearance is significantly different from KITTI. Notice the top right yellow car was flipped by Deep3DBox as it could not distinguish between the head and the tail. 

Our model SS+DR-finetune performs consistently well across these evaluations.

\subsection{Ablation Study}
 We conduct ablation studies to examine the effects of textures (T), light augmentation (LA), geometrical variations to object shapes (G), addition of distractors (D), and full randomization(T + LA + D + G). 
 
 We compare (1) LA + G + D, no textures, the object models are rendered without texture variations with a fixed color scheme, (2) T + G + D, no light augmentations, the lighting conditions are held constant, no contrast or brightness changes introduced, (3) T + LA + D: no geometrical variations, every object has fixed dimension, (4) T + LA + G: no distractors, (5) T + LA + D + G, full randomization. 
 
 Figure \ref{fig:bb_ablation} shows the comparison chart of the bounding boxes results at IoU=0.5 for the aforementioned randomization conditions for 3 datasets. Removing textures (LA + G + D) results in the most performance drop followed by removing geometrical variations for the task of object detection. 
 
 Thus, we conclude that varying object appearance and shape and size are two critical components for object detection.

\begin{figure}
\begin{center}
 \includegraphics[width=1\linewidth]{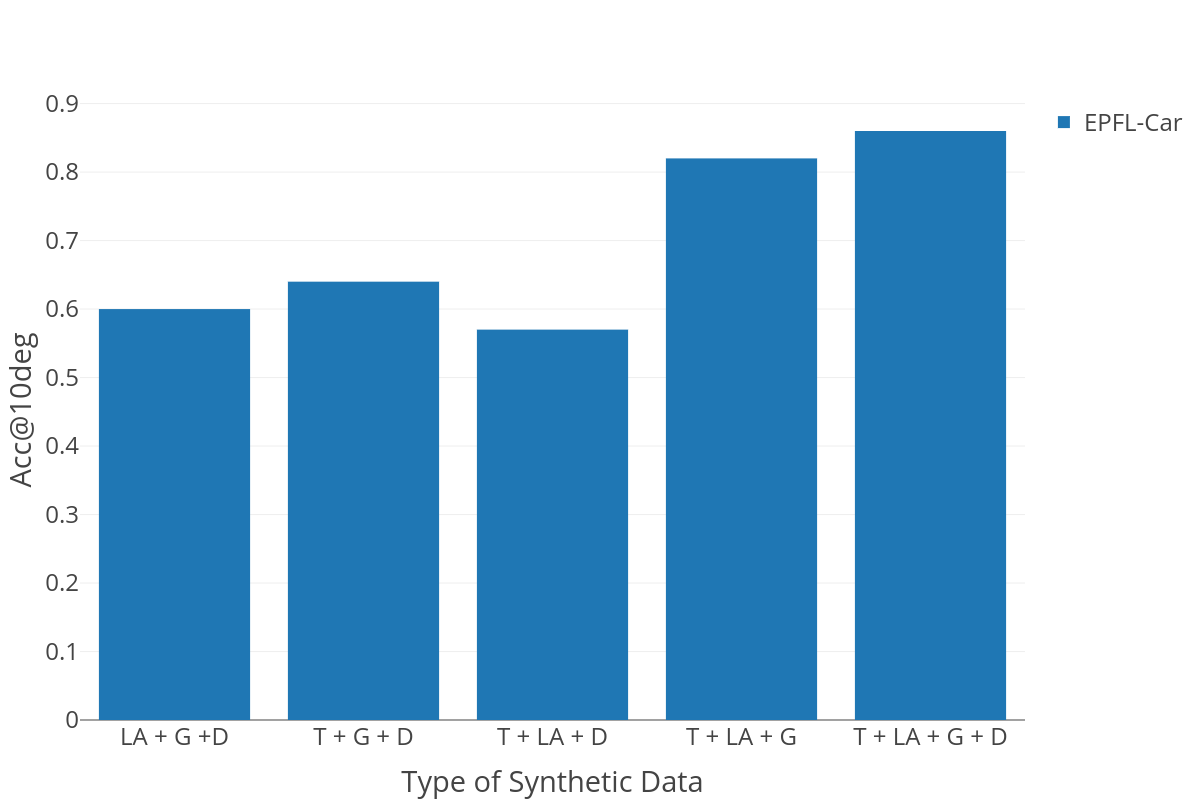}
\end{center}
   \caption{Pose ablation study on EPFL-Car dataset. Y axis is Pose Accuracy at 10 degrees, X axis various randomization conditions} 
\label{fig:pose_ablation}
\end{figure}

 Figure \ref{fig:pose_ablation} shows the comparison chart of the pose estimation results on EPFL-Cars dataset. We assume ground truth bounding box proposals. Removing geometric variations results in the most performance drop followed by the object appearance. The pose estimation head is learning appearance invariant features and is more sensitive to geometry of the object.

\section{Conclusion}
To conclude, scene specific treatment along with domain randomization is promising solution for annotation less setups in surveillance. In this paper, we proposed a framework to generate rich synthetic annotations which incorporate prior knowledge about the scene. Furthermore, using domain randomization we ensure any learning algorithm trained on such kind of data would generalize to real data during inference. We performed studies to analyse the effect of each individual randomization components. Compelling results are demonstrated on VIRAT, EPFL-Car and UA-DETRAC datasets for the detection and pose estimation. 

\section{Acknowledgement}

This work was sponsored by IARPA via Department of Interior/ Interior Business Center (D17PC00340). 

{\small
\bibliographystyle{ieee}
\bibliography{references}
}

\end{document}